\documentclass{article}

\usepackage{graphicx,spconf,subfigure}

\usepackage[normalem]{ulem}
\useunder{\uline}{\ul}{}

\newcommand\etal{{\it et~al.}}
\newcommand\ie{{\it i.e.}}

\newcommand\eg{{\it e.g.}}

\title{Scene Text Recognition with Temporal Convolutional Encoder}
\name{Xiangcheng Du$^{1,2}$,  Tianlong Ma$^{2}$, Yingbin Zheng$^{3}$, Hao Ye$^{3}$, Xingjiao Wu$^{1,2}$, Liang He$^{1,2}$}
\address{$^{1}$Shanghai Key Laboratory of Multidimensional Information Processing\\
$^{2}$East China Normal University, Shanghai, China\\
$^{3}$Videt Tech Ltd., Shanghai, China}

\begin{document}
\maketitle

\begin{abstract}
Texts from scene images typically consist of several characters and exhibit a characteristic sequence structure. Existing methods capture the structure with the sequence-to-sequence models by an encoder to have the visual representations and then a decoder to translate the features into the label sequence. In this paper, we study text recognition framework by considering the long-term temporal dependencies in the encoder stage. We demonstrate that the proposed Temporal Convolutional Encoder with increased sequential extents improves the accuracy of text recognition. We also study the impact of different attention modules in convolutional blocks for learning accurate text representations.
We conduct comparisons on seven datasets and the experiments demonstrate the effectiveness of our proposed approach.
\end{abstract}


\section{Introduction}
\label{sec:intro}

Scene text recognition is an essential task in computer vision research. The texts contain rich semantic information and are important to understand the scene images. The text recognition task has various applications such as image-based file retrieval, product recognition, search the vast information, and intelligent inspection.

Although text recognition has been studied for a few years and many approaches with promising results are proposed, recognizing text from scene images is still challenging. Recently, the deep learning based approaches treat text recognition as a sequence labeling problem, by designing an encoder to have the visual representations and then a decoder to translate the features into the label sequence. The construction of a good encoder to represent text is of fundamental importance in building a robust text recognition system. Many of previous work rely on a combination of a convolutional neural network (CNN) and a recurrent neural network (RNN) and return the representations for text sequences (\eg, in \cite{shi2016end}).

\begin{figure}[t]
  \centering
  \includegraphics[width=.7\linewidth]{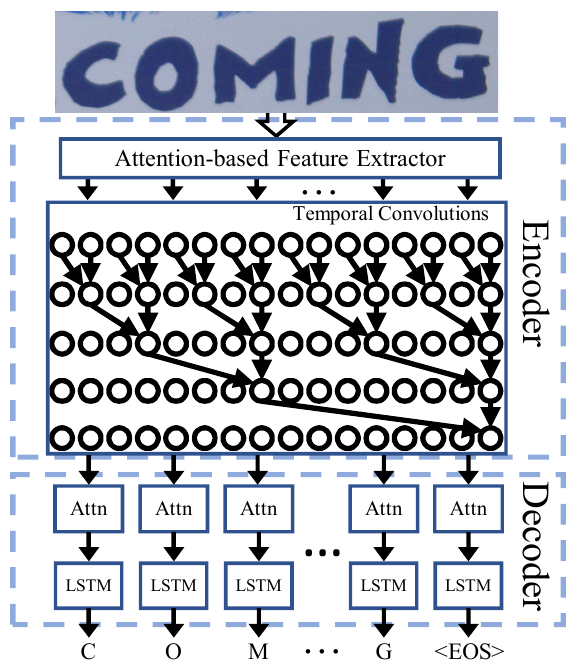}
  \caption{The network architecture.}
  \label{fig:architecture}
\end{figure}

In contrast to previous works that use RNN for sequential modeling, our framework employs the convolutional layers to capture the sequential dependencies. In this paper, we propose a new text representation called Temporal Convolutional Encoder (TCE), which is built upon the attention-based feature extractor and temporal convolutions. The utilization of long-term temporal dependencies makes TCE more discriminative than the existing works and the convolutional based design ensures the efficiency with the parallel operations. Since the accuracy of text recognition deeply depends on the abstract textual features, we also improve the CNN-based feature extractor through a novel attention mechanism.

To summarize, the highlights of this paper are:

\begin{itemize}
\item A novel Temporal Convolutional Encoder in the text recognition model by operating on the temporal resolution of the text sequences;
\item Strategies for the refinement of attention mechanisms with the channel and spatial context to improve the performance of scene text recognition;
\item Extensive evaluation on seven scene text datasets with very competitive results of the proposed encoder.
\end{itemize}

\subsection{Related Work}
\label{sec:related}

Numerous efforts have been devoted to the design of effective text representations. \cite{6945320,long2018sceneTextSurvey} surveyed the recent advances in scene text recognition approaches. In this section, we mainly discuss literature on deep learning based approaches, which are more related to this work.

Shi \etal~\cite{shi2016end} proposed an end-to-end model that captures sequence feature representation by combining CNN and RNN and then CTC loss \cite{graves2006connectionist} was used with the neural network outputs for calculating the conditional probability between the predicted and the target sequences. The CTC loss was also used in \cite{wang2017gated,he2016reading} with Gated RCNN (GRCNN) and maxout CNN. As sequence-to-sequence model, most recognition models consist of an encoder and a decoder. The attention mechanism is incorporated into the decoder. For example, Lee \etal~\cite{lee2016recursive} proposed to use an attention-based decoder for text-output prediction, while Cheng \etal~\cite{cheng2017focusing} presented Focusing Attention Network (FAN) to tackle attention drift problem in order to improve the performance of regular text recognition. Besides, some previous work also exploited to handle the irregular scene text images at the beginning of the encoder. Cheng \etal~\cite{cheng2018aon} developed arbitrary orientation network (AON) to extract features from irregular images, and \cite{shi2018aster,LUO2019109,cvpr19esir} aimed to correct image with different rectification approaches.

In this paper, the proposed approach is also towards the design of the encoder. However, instead of using a sophisticated rectifier, we adopt a spatial transformation network \cite{Jaderberg2015Spatial} to correct the image. Our focus is to construct an efficient contextual representation of the text sequences with CNNs. Probably the most related work to ours is \cite{GAO2019161}, where a fully convolutional network (FCN) was trained as the encoder. Our approach is different from \cite{GAO2019161} in its design since we use a temporal convolutional network after the attention module layers. Moreover, as will be shown in the experiments, the proposed temporal convolutional encoder outperforms the FCN based method in various benchmarks.

\section{Method}
\label{sec:method}

In this section, we will introduce our proposed framework, which consists of three subsections. We start by introducing the design of our framework, followed by the details of the attention module and temporal convolution in our encoder.

\subsection{Architecture}

We illustrate the overall architecture and networks of our framework in Fig. \ref{fig:architecture}, which are composed of two parts, \ie, the encoder and decoder. The goal of the encoder is to extract rich and discriminative visual features to represent the text regions. Classic encoder like \cite{shi2016end} employ a two-stage model: a convolutional neural network to extract the sequential feature representations from input images, and a recurrent model built upon the convolutional layers to capture contextual information. In the first stage, we observe that using attention mechanisms promotes the representation ability and thus uses a feature extractor based on ResNet \cite{2016cvpr_khe} and attention blocks. To handle the expensive computation and gradient vanishing of the recurrent model, we design the encoder with temporal convolutions to efficiently capture the long-term sequence dependencies. We will describe these components in detail in the next subsections. The decoder aims to estimate and output the text sequences from the extracted features of the text regions. While we use the attention-based prediction \cite{Baek2019what} in this paper, any decoding model will suffice (\eg, CTC~\cite{graves2006connectionist}).

\subsection{Attention-based Feature Extractor}

\begin{figure}[t]
  \centering
  \includegraphics[width=.7\linewidth]{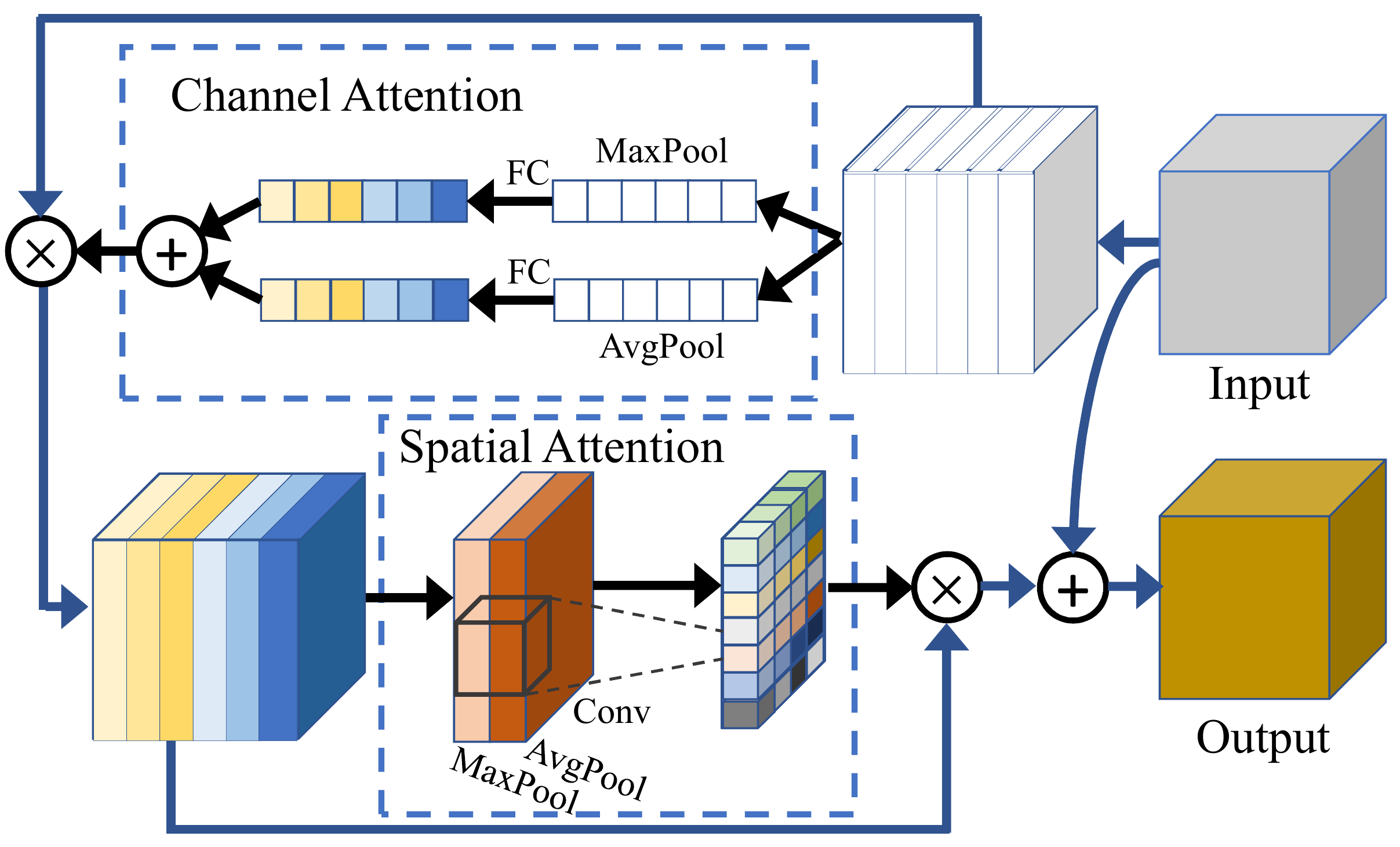}
  \caption{Pipeline of attention mechanism in feature extractor.}
  \label{fig:attention}
\end{figure}

Our baseline feature extractor employs the backbone of ResNet-18 \cite{2016cvpr_khe} for its efficiency and representation ability. Inspired by recent progress such as \cite{Hu2017SqueezeandExcitationN,woo2018cbam}, we explore variations of attention modules to enhance the discriminant of convolution layers. Fig. \ref{fig:attention} illustrates the configuration of attention for the convolution layers. We will introduce the modules as follows.

\vspace{0.08in}
\noindent\textbf{Channel attention.} Suppose we have the feature map $F$ with size of $W\times H$ and $c$ channels after the convolutional layer. The feature maps are first operated along the spatial dimension for channel attention. The average pooling and max pooling are used to have two one-dimensional vectors, \ie, $F_{\max }^c \in {R^{1 \times 1 \times c}}$ and $F_{avg}^c \in {R^{1 \times 1 \times c}}$. Then the vectors go through the full connection layer (FC) and we can obtain the channel-wise attention,
\begin{equation}
Attn_{c} = \mathrm{sigmod} \left( {\mathrm{FC}\left( {F_{avg}^c} \right) + \mathrm{FC}\left( {F_{\max }^c} \right)} \right)
\end{equation}
with multiplication with the original feature map, the transformed map is computed by ${F_{c}} = Att{n_{c}} \otimes F$.

\vspace{0.08in}
\noindent\textbf{Spatial attention}. The channel attention pays more attention to the relationship between channel features, however, it ignores the spatial information of the features. Thus, we adopt spatial attention to make up for this deficiency. The max pooling and average pooling along the channel dimension are employed to have spatial feature maps $F_{\max }^s \in {R^{H \times W \times 1}}$ and $F_{avg}^s \in {R^{H \times W \times 1}}$. A $3\times3$ convolution operation followed the concatenating of the spatial feature maps is designed for the spatial attention, \ie,
\begin{equation}
Att{n_{s}} = \mathrm{sigmod} \left( {\mathrm{conv}\left( {\mathrm{concat}\left( {F_{avg}^s,F_{\max }^s} \right)} \right)} \right)
\end{equation}
Similar with the channel attention, the result of the spatial attention is calculated as ${F_{s}} = Att{n_{s}} \otimes {F_{c}}$. To avoid the gradient vanishing with the attention modules, we also introduce the residual structure and the final transformed feature map is defined as ${F'} = F \oplus F_{s}$.

\subsection{Temporal Convolutions}

\begin{table}[t]
\centering
\caption{Setting of the temporal convolution layers. The configuration includes kernel size, channel, and dilation rate, respectively. The dropout rate is 0.3 during model training.}
\label{table:tcn}
\begin{tabular}{lcccc}
\hline
Layers & Layer1 & Layer2 & Layer3 & Layer4\\
\hline
Configuration & 3/256/1 & 3/256/2 & 3/256/4 & 3/256/8 \\
\hline
\end{tabular}
\end{table}

Sequence modeling aims to model the dependencies between feature sequences and is a key step to recognize text. As the Temporal Convolutions Network (TCN) helps the improvement on some sequence prediction tasks~\cite{lea2016temporal,bai2018empirical,wu2019fast}, here we also employ temporal convolutions to establish the relationship between feature sequences. Let ${\mathcal S}=\{S_{1},S_{2}...,S_{N}\}$ be the sequential representations generated by the extractor. To incorporate the causal relationship between the historical information, the causal convolution with one-dimension convolutional kernel $K$ is applied at the ${t}$-th element of features,
\begin{equation}
{S'_t} = \sum\limits_{i = 0}^{n - 1} {{K_i} \cdot {S_{t - i}}}
\end{equation}
where n is the size of $K$.  The length of modeling of feature sequences by the ordinary causal convolution is limited by kernel size. Capturing long-term dependencies requires more network layers or larger kernel size. However, directly increasing these values results in high computational cost.

In this paper, we use dilated convolution to increase temporal extents. The definition of dilated operation is as follows,
\begin{equation}
{S'_t}=\sum\limits_{i = 0}^{n - 1} {{K_i} \cdot {S_{t - d \times i}}}
\end{equation}
where $d$ is the expansion parameter of the dilated convolution. The dilated convolution allows for interval sampling of the input, such that the size of the receptive field grows exponentially with the number of layers. Overall, a temporal convolutional network with dilation can obtain a large receptive field with fewer layers. A diagram of our network design is shown in the middle part of Fig. \ref{fig:architecture}.

Although the dilated convolution can reduce the number of layers of the network, it still requires several network layers to obtain a complete receptive field. Meanwhile, when the channel feature information is passed between network layers, the vanishing gradient problems tend to occur. To this end, we adopt residual connections to convey feature information between network layers in sequence modeling. The network setting of our sequence model based on temporal convolutions is listed in Table \ref{table:tcn}.

\begin{table}
\centering
\caption{Statistics of the datasets. Two types are included: regular datasets (reg.) contain text images with horizontal text regions, while irregular datasets (irreg.) also contain distorted regions such as curved or rotated texts. Size indicates the testing image size of each dataset.}
\label{table:Dataset}
\begin{tabular}{@{}lcc|lcc@{}}
\hline
Dataset & Type    & Size & Dataset & Type      & Size \\ \hline
SVT     & reg. & 647      & CUTE80   & irreg. & 288      \\
ICDAR03 & reg. & 867      & SVTP    & irreg. & 645      \\
ICDAR13 & reg. & 1015     & ICDAR15 & irreg. & 1811     \\
IIIT    & reg. & 3000     &         &           &          \\ \hline
\end{tabular}
\end{table}

\begin{table*}[t]
\caption{Comparison with the state-of-the-arts on text recognition benchmarks. Bold text denotes the top result, while underlined text corresponds to the second runner-up.}
\label{table:CSTOA}
\centering
\begin{tabular}{llccccccc}
\hline
{Model}                              & Year & {SVT}         & {ICDAR03}       & {ICDAR13}       & {IIIT}        & {CUTE80}      & {SVTP}          & {ICDAR15}       \\ \hline
CRNN~\cite{shi2016end}             & 2016 & 82.7          & 91.9            & 89.6            & 81.2          & -             & -               & -               \\
{GRCNN~\cite{wang2017gated}}         & 2017 & 81.5          & 91.2            & -               & 80.8          & -             & -               & -               \\
{FAN~\cite{cheng2017focusing}}       & 2017 & {85.9}        & {94.2}          & {93.3}          & {87.4}        & {-}           & {-}             & {70.6}          \\
{AON~\cite{cheng2018aon}}            & 2018 & {82.8}        & {91.5}          & {-}             & {87.0}        & 76.8          & {73}            & {-}             \\
EP~\cite{EP}                       & 2018 & {87.5}        & 94.6            & {\textbf{94.4}} & 88.3          & {-}           & {-}             & {73.9}          \\
ASTER~\cite{shi2018aster}            & 2018 & \textbf{93.4} & 94.5            & 91.8            & \textbf{93.4} & 79.5          & 78.5            & 76.1            \\
FCN \cite{GAO2019161}                & 2019 & {82.7}        & {89.2}          & {88.0}          & {81.8}        & {-}           & {-}             & {62.3}          \\
MORAN~\cite{LUO2019109}              & 2019 & 88.3          & {\ul 95.0}      & 92.4            & 91.2          & 77.4          & 76.1            & 68.8            \\
Baek \etal~\cite{Baek2019what}     & 2019 & 87.5          & {94.4}          & {92.3}          & {87.9}        & 74.0          & {{\ul 79.2}}    & {{\ul 77.6}}    \\
SAR~\cite{aaai19sar}                 & 2019 & 84.5          & -               & 91.0            & 91.5          & \textbf{83.3} & 76.4            & 69.2            \\
ESIR~\cite{cvpr19esir}               & 2019 & {\ul 90.2}    & -               & 91.3            & {\ul 93.3}    & \textbf{83.3} & 79.6            & 76.9            \\ \hline
TCE [ours]                               &      & 89.0          & {\textbf{95.4}} & {{\ul 93.8}}    & 88.6          & {72.9}        & {\textbf{81.9}} & {\textbf{79.9}} \\ \hline
\end{tabular}
\end{table*}

\section{Experiments}
\label{sec:exp}

In this section, we will introduce the experimental setup and evaluate the performance of the proposed network.

\subsection{Configuration}

\vspace{0.08in}
\noindent\textbf{Datasets}. We evaluate the proposed text recognition network on seven standard datasets: Street View Text (SVT)~\cite{wang2011end}, SVT Perspective (SVTP)~\cite{quy2013recognizing}, IIIT~\cite{mishra2012scene}, CUTE80~\cite{risnumawan2014robust}, ICDAR03~\cite{lucas2003icdar}, ICDAR13~\cite{karatzas2013icdar}, and ICDAR15~\cite{karatzas2015icdar}. The details of the datasets are summarized in Table~\ref{table:Dataset}. Following \cite{Baek2019what}, our model is trained with the synthetic images from MJSynth~\cite{jaderberg2016reading} and SynthText~\cite{gupta2016synthetic} and the combination of training images from ICDAR13, ICDAR15, IIIT, and SVT are considered as a validation dataset. Then we can obtain a single network and apply it to the testing set of the benchmarks. Throughout the experiments, we do not employ lexicon and the mean accuracy is used to evaluate the performance.

\vspace{0.08in}
\noindent\textbf{Baselines}. We compare our temporal convolutional encoder with several baselines and state-of-the-art approaches. Among them, the first group contains several well-known recognition networks, including CRNN~\cite{shi2016end} and GRCNN~\cite{wang2017gated}. We then compare ours with previous attention aware approaches such as FAN~\cite{cheng2017focusing}, FCN~\cite{GAO2019161}, and Baek \etal~\cite{Baek2019what}. Other state-of-the-art approaches, \eg, ASTER~\cite{shi2018aster}, MORAN~\cite{LUO2019109}, and ESIR~\cite{cvpr19esir}, are also included in the comparison.

\vspace{0.08in}
\noindent\textbf{Implementation details}. Our recognition model is implemented using PyTorch~\cite{Pytorch}. Spatial transformer network \cite{Jaderberg2015Spatial} is used before the encoder to rectify the scene image. To train the network, we first resize the training images into $100\times32$ and normalize the pixel values to the range of (-1, 1). The layers of the network are initialized using a Gaussian distribution and the AdaDelta optimizer is used with the decay rate of 0.95. The training process reaches convergence after 250k iterations.

\subsection{Results and Discussion}

\vspace{0.08in}
\noindent\textbf{Comparison with the state-of-the-arts}. Table \ref{table:CSTOA} summarizes the recognition accuracies of all the benchmarks. We can see that temporal convolutional encoder outperforms all the previous methods in ICDAR03, SVTP, and ICDAR15, and is comparable with the state-of-the-arts for the rest datasets. It worth mentioning that our proposed framework can achieve better performance comparing with the FCN-based method~\cite{GAO2019161} in the benchmarks. We notice that the rectification approaches \cite{shi2018aster,cvpr19esir} also achieve good performance in all the benchmarks. As these methods and ours focus on different stages of a text recognition system, we believe that TCE is complementary to the rectification approaches and we would like to examine it in the future.

\begin{figure}[t]
  \subfigure[]{
\begin{minipage}[t]{0.6\linewidth}
\centering
\vspace{-0.74in}
\begin{tabular}{lll}
    \hline
    Dataset                       & LSTM                     & TC                               \\ \hline
    {ICDAR03}                     & 94.6                     & \textbf{95.4}                     \\ \hline
    ICDAR13 & 93.1 & \textbf{93.8} \\ \hline
    ICDAR15 & 78.1 & \textbf{79.9} \\ \hline
    \end{tabular}
    \end{minipage}
    }
    \subfigure[]{
    \begin{minipage}[t]{0.35\linewidth}
    \centering
    \includegraphics[width=\linewidth,height=.65\linewidth]{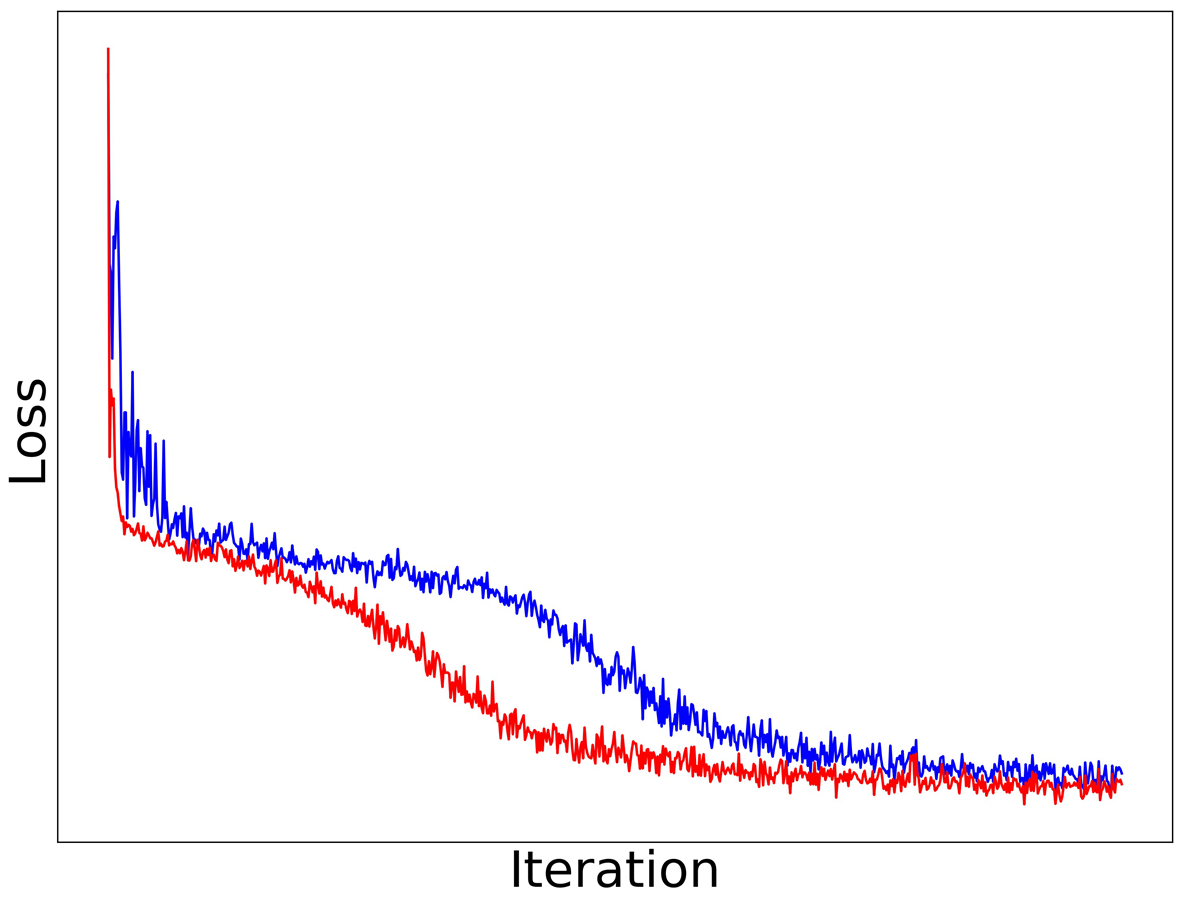}
    \end{minipage}
    }
  \caption{Experiment results (a) and training losses at different iterations on the synthetic dataset (b) with encoder using LSTM (blue curve) and temporal convolutions (TC, red curve).}
  \label{fig:errors}
\end{figure}

\begin{table}[t]
\caption{The accuracy of using different attention module. Baseline setting is with a ResNet-18 in the feature extraction stage. CA and SA are the channel attention and spatial attention, respectively.}
\label{table:attention}
\centering
\begin{tabular}{@{}lccc@{}}
\hline
Setting & ICDAR03       & ICDAR13       & ICDAR15       \\ \hline
Baseline    & 94.1          & 91.7          & 76.4          \\
CA      & {\ul 95.3}    & {\ul 92.9}    & 78.0          \\
SA      & 95.0          & 92.8          & {\ul 79.2}    \\
CA+SA   & \textbf{95.4} & \textbf{93.8} & \textbf{79.9} \\ \hline
\end{tabular}
\end{table}

\vspace{0.08in}
\noindent\textbf{Ablation study}. Here we evaluate the parameters for constructing the encoder and report results on the ICDAR datasets. We first compare the sequence modeling approaches. In Fig. \ref{fig:errors}(a), it clearly shows that using temporal convolutions can improve text recognition performance over the Bidirectional LSTM which is popular in the encoder part of previous work (\eg, in \cite{shi2018aster,Baek2019what}). We also observe that the proposed approach converges faster than the LSTM-based encoder network (see Fig. \ref{fig:errors}(b)).

We also explore the influence of different choices of attention module in the encoder. As mentioned earlier, the channel attention or spatial attention may be added by directly putting them after a convolutional layer. In Table \ref{table:attention}, we report the detailed results and find that adding attention boosts the accuracy. Moreover, using both channel and spatial attention modalities leads to performance gains for all of ICDAR datasets.

\section{Conclusions}
\label{sec:conclusion}

In this paper, we introduced a framework based on the Temporal Convolutional Encoder for scene text recognition. Text sequences were modeled with dilated convolutions to increase the temporal receptive field, resulting in a more efficient model. Channel and spatial attention mechanisms were also explored to refine the feature extractor. Experimental comparisons with the state-of-the-art approaches on seven standard datasets showed the effectiveness and efficiency of our proposed encoder for the text recognition task.

\small{
\bibliographystyle{IEEEbib}
\bibliography{arxiv}
}

\end{document}